\documentclass[10pt, conference, compsocconf]{IEEEtran}
\ifCLASSINFOpdf
\else
\fi
\usepackage{graphicx}
\usepackage{amsmath}
\usepackage{multirow}

\begin{document}
%
\title{A Deep Neural Architecture for Sentence-level Sentiment Classification in \\ Twitter Social Networking}



\author{\IEEEauthorblockN{Huy Nguyen\IEEEauthorrefmark{1} and
Minh-Le Nguyen\IEEEauthorrefmark{1}}
\IEEEauthorblockA{\IEEEauthorrefmark{1}Japan Advanced Institute of Science and Technology\\
Ishikawa, Japan \\
\{huy.nguyen, nguyenml\}@jaist.ac.jp
}}


%


\maketitle

\begin{abstract}
This paper introduces a novel deep learning framework including a lexicon-based approach for sentence-level prediction of sentiment label distribution. We propose to first apply semantic rules and then use a Deep Convolutional Neural Network (DeepCNN) for character-level embeddings in order to increase information for word-level embedding. After that, a Bidirectional Long Short-Term Memory network (Bi-LSTM) produces a sentence-wide feature representation from the word-level embedding. We evaluate our approach on three twitter sentiment classification datasets. Experimental results show that our model can improve the classification accuracy of sentence-level sentiment analysis in Twitter social networking.
\end{abstract}

\begin{IEEEkeywords}
Twitter; Sentiment classification, Opinion mining.

\end{IEEEkeywords}

%
\IEEEpeerreviewmaketitle

\section{Introduction}

Twitter sentiment classification have intensively researched in recent years \cite{go2009twitter}\cite{nakov2016semeval}. Different approaches were developed for Twitter sentiment classification by using machine learning such as Support Vector Machine (SVM) with rule-based features \cite{silva2011symbolic} and the combination of SVMs and Naive Bayes (NB) \cite{wang2012baselines}. In addition, hybrid approaches combining lexicon-based and machine learning methods also achieved high performance described in \cite{muhammad2016contextual}. However, a problem of traditional machine learning is how to define a feature extractor for a specific domain in order to extract important features.

Deep learning models are different from traditional machine learning methods in that a deep learning model does not depend on feature extractors because features are extracted during training progress. The use of deep learning methods becomes to achieve remarkable results for sentiment analysis \cite{dos2014deep}\cite{Kim2014}\cite{DBLP:journals/corr/ZhangL15}. Some researchers used Convolutional Neural Network (CNN) for sentiment classification. CNN models have been shown to be effective for NLP. For example, \cite{Kim2014} proposed various kinds of CNN to learn sentiment-bearing sentence vectors, \cite{dos2014deep} adopted two CNNs in character-level to sentence-level representation for sentiment analysis. \cite{DBLP:journals/corr/ZhangL15} constructs experiments on a character-level CNN for several large-scale datasets. In addition, Long Short-Term Memory (LSTM) is another state-of-the-art semantic composition model for sentiment classification with many variants described in \cite{gers2000recurrent}. The studies reveal that using a CNN is useful in extracting information and finding feature detectors from texts. In addition, a LSTM can be good in maintaining word order and the context of words. However, in some important aspects, the use of CNN or LSTM separately may not capture enough information.

\indent Inspired by the models above, the goal of this research is using a Deep Convolutional Neural Network (DeepCNN) to exploit the information of characters of words in order to support word-level embedding. A Bi-LSTM produces a sentence-wide feature representation based on these embeddings. The Bi-LSTM is a version of \cite{graves2013generating} with Full Gradient described in \cite{graves2005framewise}. In addition, the rules-based approach also effects classification accuracy by focusing on important sub-sentences expressing the main sentiment of a tweet while removing unnecessary parts of a tweet. The paper makes the following contributions:
\begin{itemize}
	\item We construct a tweet processor removing unnecessary sub-sentences from tweets in order that the model learns important information in a tweet. We share ideas with \cite{go2009twitter} and \cite{appel2016hybrid}, however, our tweet processor keeps emoticons in tweets and only uses rules to remove non-essential parts for handling negation.
	\item We train DeepCNN on top of character embeddings to produce feature maps which capture the morphological and shape information of a word. The morphological and shape information illustrate how words are formed, and their relationship to other words. DeepCNN transforms the character-level embeddings into global fixed-sized feature vectors at higher abstract level. Such character feature vectors contribute enriching the information of words in a sentence.
	\item  We create an integration of global fixed-size character feature vectors and word-level embedding for the Bi-LSTM. The Bi-LSTM connects the information of words in a sequence and maintains the order of words for sentence-level representation.
\end{itemize}

The organization of the present paper is as follows: In section 2, we describe the model architecture which introduces the structure of the model. We explain the basic idea of model and the way of constructing the model. Section 3 show results and analysis and section 4 summarize this paper.
\begin{figure*}
	\centering
	\includegraphics[scale=0.45]{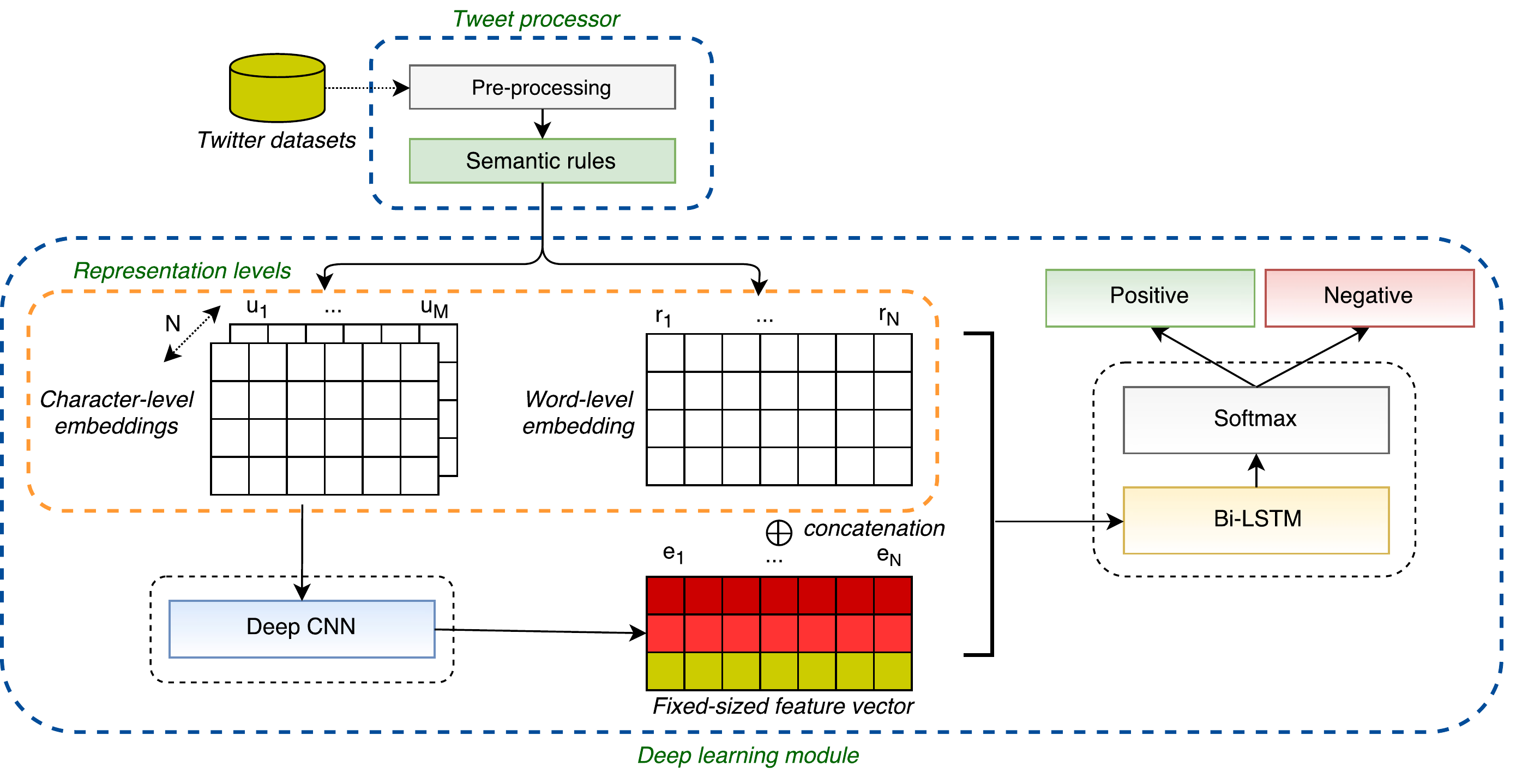}
	\caption{The overview of a deep learning system.}
	\label{model_pipeline}
\end{figure*}
\section{Model architecture}
\subsection{Basic idea}
Our proposed model consists of a deep learning classifier and a tweet processor. The deep learning classifier is a combination of DeepCNN and Bi-LSTM. The tweet processor standardizes tweets and then applies semantic rules on datasets. We construct a framework that treats the deep learning classifier and the tweet processor as two distinct components. We believe that standardizing data is an important step to achieve high accuracy. To formulate our problem in increasing the accuracy of the classifier, we illustrate our model in Figure. \ref{model_pipeline} as follows:
\begin{enumerate}
\item Tweets are firstly considered via a processor based on preprocessing steps \cite{go2009twitter} and the semantic rules-based method \cite{appel2016hybrid} in order to standardize tweets and capture only important information containing the main sentiment of a tweet.
\item We use DeepCNN with Wide convolution for character-level embeddings. A wide convolution can learn to recognize specific \textit{n-grams} at every position in a word that allows features to be extracted independently of these positions in the word. These features maintain the order and relative positions of characters. A DeepCNN is constructed by two wide convolution layers and the need of multiple wide convolution layers is widely accepted that a model constructing by multiple processing layers have the ability to learn representations of data with higher levels of abstraction \cite{lecun2015deep}. Therefore, we use DeepCNN for character-level embeddings to support morphological and shape information for a word. The DeepCNN produces $N$ global fixed-sized feature vectors for $N$ words.
\item A combination of the global fixed-size feature vectors and word-level embedding is fed into Bi-LSTM. The Bi-LSTM produces a sentence-level representation by maintaining the order of words.
\end{enumerate}

Our work is philosophically similar to \cite{dos2014deep}. However, our model is distinguished with their approaches in two aspects: 
\begin{itemize}
\item Using DeepCNN with two wide convolution layers to increase representation with multiple levels of abstraction. 
\item Integrating global character fixed-sized feature vectors with word-level embedding to extract a sentence-wide feature set via Bi-LSTM. This deals with three main problems: (i) Sentences have any different size; (ii) The semantic and the syntactic of words in a sentence are captured in order to increase information for a word; (iii) Important information of characters that can appear at any position in a word are extracted.
\end{itemize}

In sub-section B, we introduce various kinds of dataset. The modules of our model are constructed in other sub-sections.
\subsection{Data Preparation}
\begin{itemize}
	\item \textit{Stanford - Twitter Sentiment Corpus (STS Corpus):} STS Corpus contains 1,600K training tweets collected by a crawler from \cite{go2009twitter}. \cite{go2009twitter} constructed a test set manually with 177 negative and 182 positive tweets. The Stanford test set is small. However, it has been widely used in different evaluation tasks \cite{go2009twitter} \cite{dos2014deep} \cite{bravo2013combining}.
	\item \textit{Sanders - Twitter Sentiment Corpus:} This dataset consists of hand-classified tweets collected  by using search terms: $\#apple$, \textit{\#google, \#microsoft and \#twitter}. We construct the dataset as \cite{da2014tweet} for binary classification.
	\item \textit{Health Care Reform (HCR):} This dataset was constructed by crawling tweets containing the hashtag \textit{\#hcr} \cite{speriosu2011twitter}. Task is to predict positive/negative tweets \cite{da2014tweet}.

\end{itemize}
\begin{table*}
	\centering
	\caption{Semantic rules \cite{appel2016hybrid}}
	\label{semantic_rules}
	\begin{tabular}{c|p{8cm}|p{5cm}|p{2cm}}
		\hline
		Rule & Semantic rules & Example - STS Corpus & Output\\\hline
		R11 & If a sentence contains "but", disregard all previous sentiment and only take the sentiment of the part after "but". & @kirstiealley my dentist is great \textit{but} she's expensive...=( &  she's expensive...=( \\\hline
		R12 & If a sentence contains "despite", only take sentiment of the part before "despite". & I'm not dead \textit{despite} rumours to the contrary. & I'm not dead   \\\hline
		R13 & If a sentence contains "unless", and "unless" is followed by a negative clause, disregard the "unless" clause. & laptop charger is broken - \textit{unless} a little cricket set up home inside it overnight. typical at the worst possible time. & laptop charger is broken \\\hline
		R14 & If a sentence contains "while", disregard the sentence following the "while" and take the sentiment only of the sentence that follows the one after the "while". & My throat is killing me, and \textit{While} I got a decent night's sleep last night, I still feel like I'm about to fall over.  &  I still feel like I'm about to fall over \\\hline
		R15 & If the sentence contains "however", disregard the sentence preceding the "however" and take the sentiment only of the sentence that follows the "however". & @lonedog bwahahah...you are amazing!  \textit{However}, it was quite the letdown. & it was quite the letdown. \\\hline
	\end{tabular}
\end{table*}
    \begin{table}
	\centering
	\caption{The number of tweets are processed by using semantic rules}
	\label{processed_tweets}
	\begin{tabular}{|l|c|c|}
		\hline
		\textbf{Dataset} & \textbf{Set} & \textbf{\# Sentences/ tweets} \\\hline
		\multirow{2}{*}{STS Corpus}  & Train & 138703 \\
		& Test& 25 
		\\\hline
		\multirow{3}{*}{Sanders} & Train & 39 \\
		& Dev & 74 \\ 
		& Test & 54 
		\\\hline
		HCR & Train & 164
		\\\hline
	\end{tabular}
\end{table}
\subsection{Preprocessing}
We firstly take unique properties of Twitter in order to reduce the feature space such as \textit{Username}, \textit{Usage of links}, \textit{None}, \textit{URLs} and \textit{Repeated Letters}. We then process \textit{retweets}, \textit{stop words}, \textit{links}, \textit{URLs}, \textit{mentions}, \textit{punctuation} and \textit{accentuation}. For emoticons, \cite{go2009twitter} revealed that the training process makes the use of emoticons as noisy labels and they stripped the emoticons out from their training dataset because \cite{go2009twitter} believed that if we consider the emoticons, there is a negative impact on the accuracies of classifiers. In addition, removing emoticons makes the classifiers learns from other features (e.g. unigrams and bi-grams) presented in tweets and the classifiers only use these non-emoticon features to predict the sentiment of tweets. However, there is a problem is that if the test set contains emoticons, they do not influence the classifiers because emoticon features do not contain in its training data. This is a limitation of \cite{go2009twitter}, because the emoticon features would be useful when classifying test data. Therefore, we keep emoticon features in the datasets because deep learning models can capture more information from emoticon features for increasing classification accuracy.

\subsection{Semantic Rules (SR)}
In Twitter social networking, people express their opinions containing sub-sentences. These sub-sentences using specific PoS particles (Conjunction and Conjunctive adverbs), like \textit{"but, while, however, despite, however"} have different polarities. However, the overall sentiment of tweets often focus on certain sub-sentences. For example: 
\begin{itemize}
\item \textit{@lonedog bwahahah...you are amazing! However, it was quite the letdown.}
\item \textit{@kirstiealley my dentist is great but she's expensive...=(}
\end{itemize}

In two tweets above, the overall sentiment is negative. However, the main sentiment is only in the sub-sentences following \textit{but} and \textit{however}. This inspires a processing step to remove unessential parts in a tweet.  
Rule-based approach can assists these problems in handling negation and dealing with specific PoS particles led to effectively affect the final output of classification \cite{appel2016hybrid}\cite{xie2014muses}. \cite{appel2016hybrid} summarized a full presentation of their semantic rules approach and devised ten semantic rules in their hybrid approach based on the presentation of \cite{xie2014muses}. We use five rules in the semantic rules set because other five rules are only used to compute polarity of words after POS tagging or Parsing steps. We follow the same naming convention for rules utilized by \cite{appel2016hybrid} to represent the rules utilized in our proposed method. The rules utilized in the proposed method are displayed in Table \ref{semantic_rules} in which is included examples from STS Corpus and output after using the rules. Table \ref{processed_tweets} illustrates the number of processed sentences  on each dataset.

\subsection{Representation Levels}
\begin{figure*}
	\centering
	\includegraphics[scale=0.4]{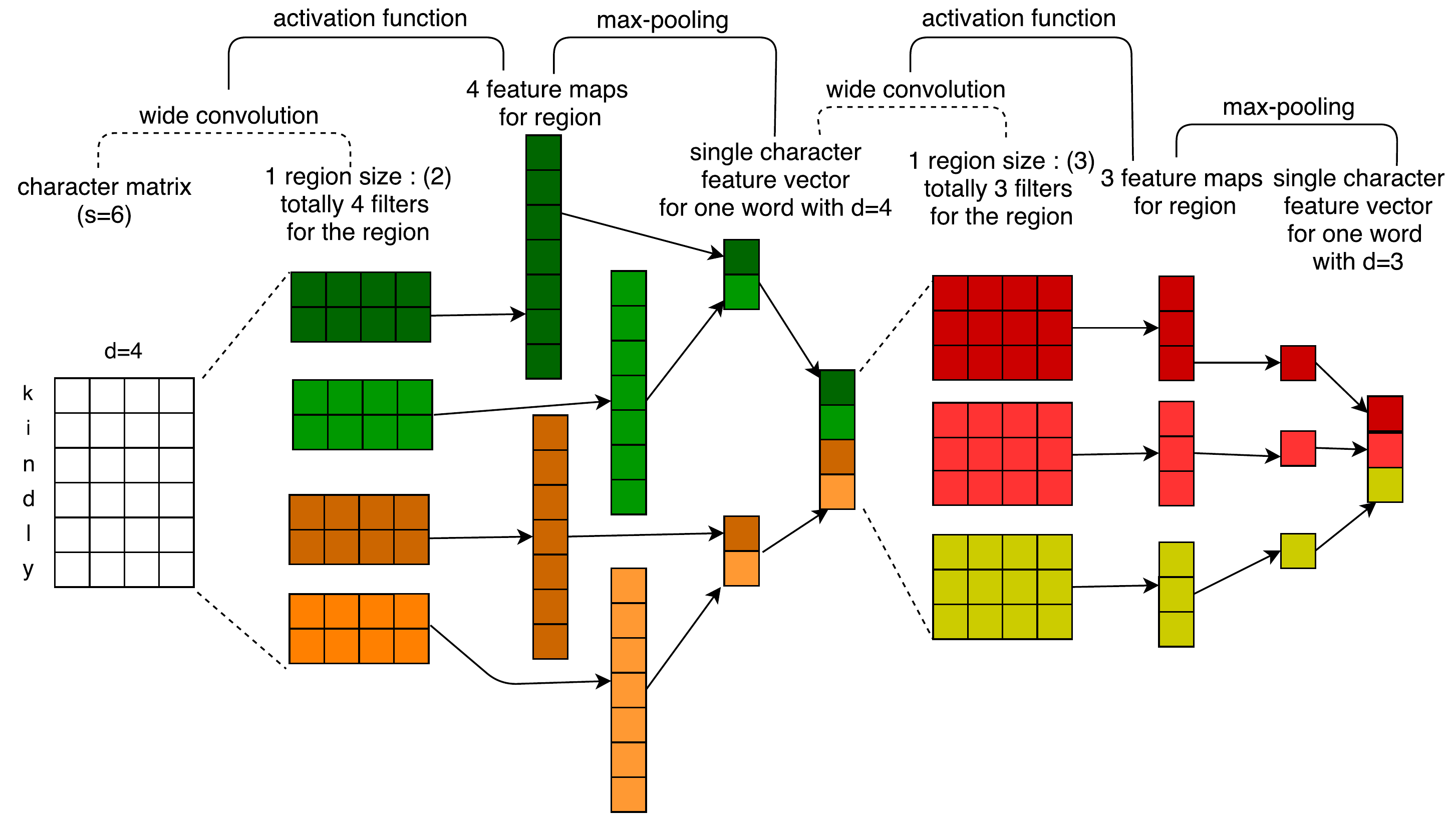}
	\caption{Deep Convolutional Neural Network (DeepCNN) for the sequence of character embeddings of a word. For example with 1 region size is 2 and 4 feature maps in the first convolution and 1 region size is 3 with 3 feature maps in the second convolution.}
	\label{deep_convolution}
\end{figure*}
To construct embedding inputs for our model, we use a fixed-sized word vocabulary $V^{word}$ and a fixed-sized character vocabulary $V^{char}$. Given a word $w_i$ is composed from characters $\{c_1, c_2, ..., c_M\}$, the character-level embeddings are encoded by column vectors $u_i$ in the embedding matrix $W^{char} \in \!R^{d^{char} \times |V^{char}|}$, where $V^{char}$ is the size of the character vocabulary.  For word-level embedding $r_{word}$, we use a pre-trained word-level embedding with dimension 200 or 300. A pre-trained word-level embedding can capture the syntactic and semantic information of words \cite{NIPS2013_5021}.  We build every word $w_i$ into an embedding $v_i = [r_{i}; e_{i}]$ which is constructed by two sub-vectors: the word-level embedding $r_{i} \in \!R^{d^{word}}$ and the character fixed-size feature vector $e_{i} \in \!R^l$ of $w_i$ where $l$ is the length of the filter of wide convolutions. We have $N$ character fixed-size feature vectors corresponding to word-level embedding in a sentence.
\subsection{Deep Learning Module}
DeepCNN in the deep learning module is illustrated in Figure. \ref{deep_convolution}. The DeepCNN has two wide convolution layers. The first layer extract local features around each character windows of the given word and using a max pooling over character windows to produce a global fixed-sized feature vector for the word. The second layer retrieves important context characters and transforms the representation at previous level into a representation at higher abstract level. We have $N$ global character fixed-sized feature vectors for $N$ words.

In the next step of Figure. \ref{model_pipeline}, we construct the vector $v_i = [r_i, e_i]$ by concatenating the word-level embedding with the global character fixed-size feature vectors. The input of Bi-LSTM is a sequence of embeddings $\{v_1, v_2, ..., v_N\}$. The use of the global character fixed-size feature vectors increases the relationship of words in the word-level embedding. The purpose of this Bi-LSTM is to capture the context of words in a sentence and maintain the order of words toward to extract sentence-level representation. The top of the model is a softmax function to predict sentiment label. We describe in detail the kinds of CNN and LSTM that we use in next sub-part 1 and 2.
\subsubsection{Convolutional Neural Network}
The one-dimensional convolution called time-delay neural net has a filter vector $m$ and take the dot product of filter $m$ with each \textit{m-grams} in the sequence of characters $s_i \in \!R$ of a word in order to obtain a sequence $c$:
\begin{equation}
	c_j = m^Ts_{j-m+1:j}
\end{equation}
Based on Equation 1, we have two types of convolutions that depend on the range of the index $j$. The narrow type of convolution requires that $s \geq m$ and produce a sequence $c \in \!R^{s-m+1}$. The wide type of convolution does not require on $s$ or $m$ and produce a sequence $c \in \!R^{s+m-1}$. Out-of-range input values $s_i$ where $i < 1$ or $i > s$ are taken to be zero. We use wide convolution for our model.
\subsubsection*{Wide Convolution}
Given a word $w_i$ composed of $M$ characters $\{c_1, c_2, ..., c_M\}$, we take a character embedding $u_i \in \!R^d$ for each character $c_i$ and construct a character matrix $W^{char} \in \!R^{d \times |V^{chr}|}$ as following Equation. 2:
\begin{equation}
  W^{char} = \begin{bmatrix}
		| & | & | \\
		u_1 & ... & u_M \\
		| & | & |
	\end{bmatrix}
\end{equation}
The values of the embeddings $u_i$ are parameters that are optimized during training. The trained weights in the filter $m$ correspond to a feature detector which learns to recognize a specific class of \textit{n-grams}. The \textit{n-grams} have size $n \geq m$. The use of a wide convolution has some advantages more than a narrow convolution because a wide convolution ensures that all weights of filter reach the whole characters of a word at the margins. The resulting matrix has dimension $d \times (s + m -1)$. 

\subsubsection{Long Short-Term Memory}
Long Short-Term Memory networks usually called LSTMs are a improved version of RNN. The core idea behind LSTMs is the cell state which can maintain its state over time, and non-linear gating units which regulate the information flow into and out of the cell. The LSTM architecture that we used in our proposed model is described in \cite{graves2013generating}. A single LSTM memory cell is implemented by the following composite function:
\begin{equation}
	i_t = \sigma(W_{xi} x_t + W_{hi} h_{t-1} + W_{ci} c_{t - 1} + b_i)
\end{equation}
\begin{equation}
	f_t = \sigma(W_{xf} x_t + W_{hf} h_{t - 1} + W_{cf} c_{t - 1} + b_f )
\end{equation}
\begin{equation}
	c_t = f_t c_{t - 1} + i_t tanh(W_{xc} x_t + W_{hc} h_{t - 1} + b_c)
\end{equation}
\begin{equation}
	o_t = \sigma(W_{xo} x_t + W_{ho} h_{t - 1} + W_{co} c_t + b_o)
\end{equation}
\begin{equation}
	h_t = o_t tanh(c_t)
\end{equation}
where $\sigma$ is the logistic sigmoid function, $i, f, o$ and $c$ are the \textit{input gate, forget gate, output gate, cell} and \textit{cell input} activation vectors respectively. All of them have a same size as the hidden vector $h$.  $W_{hi}$ is the hidden-input gate matrix, $W_{xo}$ is the input-output gate matrix. The bias terms which are added to $i, f, c$ and $o$ have been omitted for clarity. In addition, we also use the full gradient for calculating with full backpropagation through time (BPTT) described in \cite{graves2005framewise}. A LSTM gradients using finite differences could be checked and making practical implementations more reliable.
\subsection{Regularization}
For regularization, we use a constraint on $l_2-norms$ of the weight vectors \cite{hinton2012improving}.

\section{Results and Analysis}
\subsection{ Experimental setups}
For the Stanford Twitter Sentiment Corpus, we use the number of samples as \cite{dos2014deep}. The training data is selected 80K tweets for a training data and 16K tweets for the development set randomly from the training data of \cite{go2009twitter}. We conduct a binary prediction for STS Corpus. \\
\indent For Sander dataset, we use standard 10-fold cross validation as \cite{da2014tweet}. We construct the development set by selecting 10\% randomly from 9-fold training data. 

In Health Care Reform Corpus, we also select 10\% randomly for the development set in a training set and construct as \cite{da2014tweet} for comparison. We describe the summary of datasets in Table III.
\begin{table}[h]
	\centering
	\caption{Summary statistics for the datasets after using semantic rules. $c$: the number of classes. $N$: The number of tweets. $l_{w}$: Maximum sentence length. $l_c$: Maximum character length. $|V_w|$: Word alphabet size. $|V_c|$: Character alphabet size.}
	\begin{tabular}{c|c|c|c|c|c|c|c|c}
		\hline
		\textbf{Data} & \textbf{ Set} & \textit{N}	&	\textit{c} & $l_{w}$	&	$l_{c}$	& $|V_{w}|$  & $|V_{c}|$ & \textit{Test} \\\hline
		\multirow{3}{*}{STS}		&	Train	&	80K		& 	\multirow{3}{*}{2}	& 	33	&	110		&	\multirow{3}{*}{67083}	&	\multirow{3}{*}{134} & 	\multirow{3}{*}{-} 	 \\
		&	Dev		&	16K		&								&	28	&	48	&										&										\\
		&	Test	&	359		&								&	21	&	16	&										&										\\\hline
		\multirow{3}{*}{Sanders}	&	Train	&	991		& 	\multirow{3}{*}{2}	& 	31	&	33		&	\multirow{3}{*}{3379}	&	\multirow{3}{*}{84}	& 	\multirow{3}{*}{CV}	 \\
		&	Dev		&	110		&									&	27	&	47	&										&										\\
		&	Test	&	122		&									&	28	&	21	&										&										\\\hline
		\multirow{3}{*}{HCR}	&	Train	&	621		& 	\multirow{3}{*}{2}	& 	25	&	70		&	\multirow{3}{*}{3100}	&	\multirow{3}{*}{60}	& 	\multirow{3}{*}{-}	 \\
		&	Dev		&	636		&									&	26	&	16	&										&										\\
		&	Test	&	665		&									&	20	&	16	&										&										\\\hline
	\end{tabular}
\end{table}
\subsubsection{Hyperparameters}
for all datasets, the filter window size ($h$) is 7 with 6 feature maps each for the first wide convolution layer, the second wide convolution layer has a filter window size of 5 with 14 feature maps each. Dropout rate ($p$) is 0.5, $l_2$ constraint, learning rate is 0.1 and momentum of 0.9. Mini-batch size for STS Corpus is 100 and others are 4. In addition, training is done through stochastic gradient descent over shuffled mini-batches with Adadelta update rule \cite{zeiler2012adadelta}.
\subsubsection{Pre-trained Word Vectors}
we use the publicly available \textit{Word2Vec} trained from 100 billion words from Google and \textit{TwitterGlove}\footnote{https://nlp.stanford.edu/projects/glove/} of Stanford is performed on aggregated global word-word co-occurrence statistics from a corpus. \textit{Word2Vec} has dimensionality of 300 and \textit{Twitter Glove} have dimensionality of 200. Words that do not present in the set of pre-train words are initialized randomly.
\begin{table}[h]
	\centering
    \caption{Accuracy of different models for binary classification}
	\begin{tabular}{p{4.5cm}|c|c|c}
		\hline
		\textbf{Model} & \textbf{STS} & \textbf{Sanders} & \textbf{HCR} \\\hline
		CharSCNN/ pre-training \cite{dos2014deep} & \textbf{86.4} & - & - \\
		CharSCNN/ random \cite{dos2014deep} & 81.9 & - & - \\
		SCNN/ pre-training \cite{dos2014deep} & 85.2  & - & - \\
		SCNN/ random \cite{dos2014deep} & 82.2 & - & -\\
		MaxEnt \cite{go2009twitter} & 83.0 & - & -\\
		NB \cite{go2009twitter} & 82.7 & - & -\\
		SVM \cite{go2009twitter} & 82.2 & - & - \\ 
		SVM-BoW & - & 82.43 & 73.99 \\
		SVM-BoW + lex & - & 83.98 & 75.94 \\
		RF-BoW & - & 79.24 & 70.83 \\
		RF-BoW + lex & - & 82.35 & 72.93 \\
		LR-BoW & - & 77.45 & 73.83 \\
		LR-BoW + lex & - & 79.49 & 74.73 \\
		MNB-BoW & - & 79.82 & 72.48 \\
		MNB-BoW + lex & - & 83.41 & 75.33 \\
		ENS (RF + MNB + LR) - BoW & - & - & 75.19 \\
		ENS (SVM + RF + MNB + LR) - BoW + lex & - & - & \textbf{76.99} \\
		
		ENS (SVM + RF + MNB + LR) - BoW & - & 82.76 & - \\
		ENS (SVM + RF + MNB) - BoW + lex & - & \textbf{84.89} & - \\ \hline
        DeepCNN + SR + Glove & 85.23 & 62.38 & 76.84 \\ 
        Bi-LSTM + SR + Glove & 85.79 & 84.32 & 78.49 \\
        (DeepCNN + Bi-LSTM) + SR + Glove & \textbf{86.63} & \textbf{85.14} & 79.55 \\
        (DeepCNN + Bi-LSTM) + SR + GoogleW2V & 86.35 & 85.05 & \textbf{80.9} \\
        (DeepCNN + Bi-LSTM) + GoogleW2V & 86.07 & 84.23 & 80.75
		\\\hline
	\end{tabular}	
\end{table}
\subsection{Experimental results}
Table IV shows the result of our model for sentiment classification against other models.  We compare our model performance with the approaches of  \cite{go2009twitter}\cite{dos2014deep} on STS Corpus. \cite{go2009twitter} reported the results of Maximum Entropy (MaxEnt), NB, SVM on STS Corpus having good performance in previous time. The model of \cite{dos2014deep} is a state-of-the-art so far by using a CharSCNN. As can be seen, \textit{86.63} is the best prediction accuracy of our model so far for the STS Corpus.  

For Sanders and HCR datasets, we compare results with the model of \cite{da2014tweet} that used a ensemble of multiple base classifiers (ENS) such as NB, Random Forest (RF), SVM and Logistic Regression (LR). The ENS model is combined with bag-of-words (BoW), feature hashing (FH) and lexicons. The model of \cite{da2014tweet} is a state-of-the-art on Sanders and HCR datasets. Our models outperform the model of \cite{da2014tweet} for the Sanders dataset and HCR dataset. 
\subsection{Analysis}
As can be seen, the models with SR outperforms the model with no SR. Semantic rules is effective in order to increase classification accuracy. We evaluate the efficiency of SR for the model in Table V of our full paper \footnote{https://github.com/huynt-plus/papers/blob/master/deepnn-full-paper.pdf}. We also conduct two experiments on two separate models: DeepCNN and Bi-LSTM in order to show the effectiveness of combination of DeepCNN and Bi-LSTM. In addition, the model using \textit{TwitterGlove} outperform the model using \textit{GoogleW2V} because \textit{TwitterGlove} captures more information in Twitter than \textit{GoogleW2V}. These results show that the character-level information and SR have a great impact on Twitter Data. The pre-train word vectors are good, universal feature extractors. The difference between our model and other approaches is the ability of our model to capture important features by using SR and combine these features at high benefit. The use of DeepCNN can learn a representation of words in higher abstract level. 
The combination of global character fixed-sized feature vectors and a word embedding helps the model to find important detectors for particles such as 'not' that negate sentiment and potentiate sentiment such as 'too', 'so' standing beside expected features. The model not only learns to recognize single n-grams, but also patterns in n-grams lead to form a structure significance of a sentence.


\section{Conclusions}
In the present work, we have pointed out that the use of character embeddings through a DeepCNN to enhance information for word embeddings built on top of \textit{Word2Vec} or \textit{TwitterGlove} improves classification accuracy in Tweet sentiment classification. Our results add to the well-establish evidence that character vectors are an important ingredient for word-level in deep learning for NLP. In addition, semantic rules contribute handling non-essential sub-tweets in order to improve classification accuracy.



%



\bibliography{references}
\bibliographystyle{IEEEtran}

\end{document}